\documentclass[10pt,twocolumn,letterpaper]{article}

\usepackage[breaklinks=true,bookmarks=false]{hyperref}
\usepackage{iccv}
\usepackage{times}
\usepackage{epsfig}
\usepackage{graphicx}
\usepackage{amsmath}
\usepackage{amssymb}
\usepackage{iccv}
\usepackage{times}
\usepackage{epsfig}
\usepackage{graphicx}
\usepackage{amsmath}
\usepackage{amssymb}
\usepackage{epstopdf}
\usepackage{algorithm}
\usepackage{amsmath}
\usepackage{amssymb}
\usepackage{eso-pic}
\usepackage{algorithmicx}
\usepackage{lineno}
\usepackage{subfigure}
\usepackage{algpseudocode}
\usepackage{authblk}
\usepackage{caption}
\usepackage{caption3}
\usepackage[breaklinks=true,bookmarks=false]{hyperref}

\iccvfinalcopy 
\begin{document}

\title{Fast 3D Salient Region Detection in Medical Images using GPUs}

\author{Rahul Thota \thanks{Currently with vizury.com}, Sharan Vaswani \thanks{Currently with the Department of Computer Science, University of British Columbia}, Amit Kale and Nagavijayalakshmi Vydyanathan\\
Imaging and Computer Vision Group, \\
Siemens Corporate Research and Technology,\\
Prestige Alecto Building, Electronics City,\\
Bangalore 560100, India\\
{\tt\small kale.amit@siemens.com}
}

\maketitle

\begin{abstract}
Automated detection of visually salient regions is an active area of research in computer vision. Salient regions can serve as inputs for object detectors as well as inputs for region-based registration algorithms. In this paper we consider the problem of speeding up computationally intensive bottom-up salient region detection in 3D medical volumes. The method uses the Kadir-Brady formulation of saliency. We show that in the vicinity of a salient region, entropy is a monotonically increasing function of the degree of overlap of a candidate window with the salient region. This allows us to initialize a sparse seed-point grid as the set of tentative salient region centers and iteratively converge to the local entropy maxima, thereby reducing the computation complexity compared to the Kadir Brady approach of performing this computation at every point in the image. We propose two different approaches for achieving this. The first approach involves evaluating entropy in the four quadrants around the seed point and iteratively moving in the direction that increases entropy. The second approach we propose makes use of mean shift tracking framework to affect entropy maximizing moves.
Specifically, we propose the use of uniform pmf as the target distribution to seek high entropy regions. We demonstrate the use of our algorithm on medical volumes for left ventricle detection in PET images and tumor localization in brain MR sequences.
\end{abstract}
\vspace{-0.7cm}
\section{Introduction}
Images contain great amounts of information at multiple frequencies, scales and spatial locations. Perceiving this information in its entirety is a hard task. The human brain has evolved neurological processes that selectively focus attention ~\cite{FrintropSurvey} at one salient frequency, scale or image location at a time.  In order to enable processing of large amounts of data, it is natural to automate these neurological processes. This has led to the development of automated visual saliency detection algorithms that estimate salient features in images ~\cite{ittiKoch,ittiKoch2}. Frintrop et al.\ in ~\cite{frintrop08} showed the higher repeatability and discriminative power of salient regions as compared to ensemble-based detections, leading to more accurate matching across different scenes for registration or 3D scene estimation. Object detection and recognition accuracies have been improved by applying saliency filter as the front end followed by specific descriptors like SIFT ~\cite{SIFT} trained on the object class. \\
%
Visual saliency, can be thought of as a combination of bottom-up and top-down attention ~\cite{DesimoneandDuncan1995} ~\cite{Torralba}
Top-down uses the prior knowledge, models and abstractions ~\cite{CorbettaandSchulman2002}. 
 Judd et al ~\cite{judd2009learning} used a combination of low level local features and semantic information from high level detectors for faces and people. Such top down approaches try to predict the way humans perceive the visual world. It is difficult to translate such an approach to domains like medical image analysis where human attention is task dependent. Hence, we primarily concentrate on bottom-up saliency, which predominantly depends on the conspicuity emerging from contrasting/distinguishing local image features.
For instance, Mahadevan and Vasconselos ~\cite{vijay} proposed an architecture where saliency is proportional to the discriminability of a set of features that classify center from surround, at that location. They set the size of the center window to a value comparable to the size of the display items and the surround window six times the size. This ratio is motivated from the neurological evidences on natural images. However, such an assumption does not hold in medical images where the lesions and organs can take diverse range of sizes. Itti and Koch ~\cite{ittiKoch} deal with this problem by estimating the size of salient region by applying difference of Gaussians on the image pyramid using different combinations of standard deviations that capture different center-surround ratios. This adds an extra dimension of complexity leading to high compute expense. ~\cite{le2012visual} estimated the conditional entropy of the center given its surrounds using kd-tree to reduce computation. This method assumes fixed size windows for defining center and surround region making it intractable for medical imaging where anatomies could be of various sizes. Also their method is not scalable for finding salient regions in 3 dimensional volumes.
Considering the above mentioned shortcomings we adopt the saliency definition proposed by Kadir and Brady ~\cite{kadirbrady}. Specifically, their framework considers different sized neighborhoods of every point in an image. The maximum of the product of the local entropy and the rate of the pdf as a function of scale of the neighborhood is then computed. If this maxima exceeds a pre-decided threshold the point is designated as a salient point. Subsequently Expectation-Maximization (EM) based clustering is employed to coalesce nearby detections. These steps of computing entropy and differential pdf at every point in the image at multiple scales are computationally intensive, especially when considering 3D volumetric imagery common in medical imaging.
In this paper we propose an approach for fast salient region detection in 3D imagery based on the Kadir-Brady approach. We show that in the vicinity of a salient region, entropy is a monotonically increasing function of the degree of overlap of a candidate window with the salient region. This allows us to initialize a sparse seed-point grid as the set of tentative salient region centers and iteratively converge to the local entropy maxima. This reduces the computation considerably compared to the Kadir Brady approach of performing this computation at every point in the image. We propose two different approaches for achieving this. The first approach involves evaluating entropy in the four quadrants around the seed point and iteratively moving in the direction that increases entropy. In particular, the effective displacement is calculated as the summation of four quadrant displacements weighted by corresponding normalized entropies. The second approach we propose makes use of pixel level information in a mean shift tracking framework, to effect entropy maximizing moves. Specifically, we propose the use of uniform pmf as the target distribution to seek high entropy regions. We also extend this Saliency shift algorithm to capture 3D salient regions in medical volumes by estimating orientation using 3D extension of ABMSOD algorithm ~\cite{ABMSOD}. We develop an optimized GPU implementation of the saliency seek algorithm to enable accelerate the detection of salient regions. We demonstrate results for Left ventricle detection in PET and for the tumor map in brain MR sequence. 

\vspace{-0.2cm}
\section{Technical details}
\subsection{Motivation}
\begin{figure}[ht]
\centering
\includegraphics[width=0.45\linewidth]{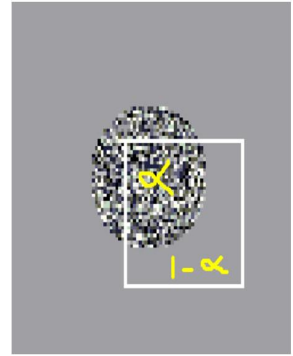}
\caption{Toy example to demonstrate monotonicity of entropy against percentage overlap with the salient region. The figure shows a target object with a uniform range of intensities in a homogeneous background region. The candidate window partially overlaps with the target with the fraction of overlap given by $\alpha$}
\label{fig:SyntheticToy}
\end{figure}

As discussed before, the Kadir Brady approach considers different sized neighborhoods of every point in an image. The maximum of the product of the local entropy and the rate of the pdf as a function of scale of the neighborhood is then computed. If this maxima exceeds a pre-decided threshold the point is designated as a salient point. Subsequently Expectation-Maximization (EM) based clustering is employed to coalesce nearby detections. The key problem with this approach is the need to perform this computation at every point in the image which as we show can be quite unnecessary. 
To motivate this, consider a toy example consisting of a single salient region as shown in the figure ~\ref{fig:SyntheticToy}. 

We assume that the intensity distribution $f$ in the salient region follows a uniform distribution and the background distribution $g$ follows a delta distribution.
Specifically $f \sim U[0,M]$ and $g \sim \delta{(k-l)}$. The choice of these distributions capture the fact that salient regions have higher entropy as compared to their background in an exaggerated sense. Now consider a candidate window with an overlap fraction $\alpha$ with the salient region. 
It is easy to see that the intensity distribution of this window will be a mixture distribution $\alpha f + (1-\alpha) g$. The entropy of this mixture distribution 
is given by: 

{\footnotesize 
\begin{equation} 
\label{eq:entropy}
H(\alpha) = -\sum_{k=0}^{M}{(\alpha f + (1-\alpha)g)\log{(\alpha f + (1-\alpha)g)}}
\end{equation}
}

Substituting the distributions for $f$ and $g$ from above, the entropy evaluates to 
{\footnotesize 
\begin{equation} 
\label{eq:entropy1}
H(\alpha) = -(\frac{\alpha}{M} + 1-\alpha)\log{(\frac{\alpha}{M} + 1-\alpha)}  -  \alpha \frac{M-1}{M} \log{\frac{\alpha}{M}}
\end{equation}
}
The rate of change of this entropy as a function of $\alpha$ is given by ~\ref{eq:entropyderi}
{\footnotesize 
\begin{equation}
\label{eq:entropyderi}
\frac{d}{d\alpha}(H(\alpha)) = \frac{M-1}{M}\log{\frac{\alpha + M(1-\alpha)}{\alpha}}
\end{equation}
}

Note that for $\alpha = 0$ $H(\alpha)=0$ and for $\alpha = 1$, $H(\alpha)=log(M)$. Also, it can be seen that the above expression in \ref{eq:entropyderi} for the derivative of the entropy is positive, for $\alpha$ ranging between 0 and 1. 
This shows that for the simple toy example that we have considered, that entropy increases monotonically as a function of overlap percentage in the vicinity of the salient object. In turn, this relationship suggests that, in the neighborhood of a salient region, an iterative algorithm that increases entropy of the enclosed region, will increase the overlap percentage. This would result in moving the candidate window closer to the salient object, obviating the need to perform an exhaustive entropy computation at every point.

In the following sections we describe two approaches which use the above idea to detect salient regions starting from a sparse set of initial seed points. The first approach involves evaluating entropy in the four quadrants centered around the seed point and iteratively moving in the direction that increases entropy. In particular, the effective displacement is calculated as the summation of four quadrant displacements weighted by corresponding normalized entropies. 
The second approach we propose makes use of mean shift tracking framework to effect entropy maximizing moves.
Specifically, we propose the use of uniform pmf as the target distribution to seek high entropy regions.
We would like to note here that while the above assumptions made regarding the distributions of the salient region and its background do not strictly hold for real world problems, the proposed entropy maximizing algorithms, nevertheless succeed in converging to the salient regions.
\subsection{Quadrant method}
\begin{figure}[ht]
\subfigure[]{
\includegraphics[width=0.55\linewidth]{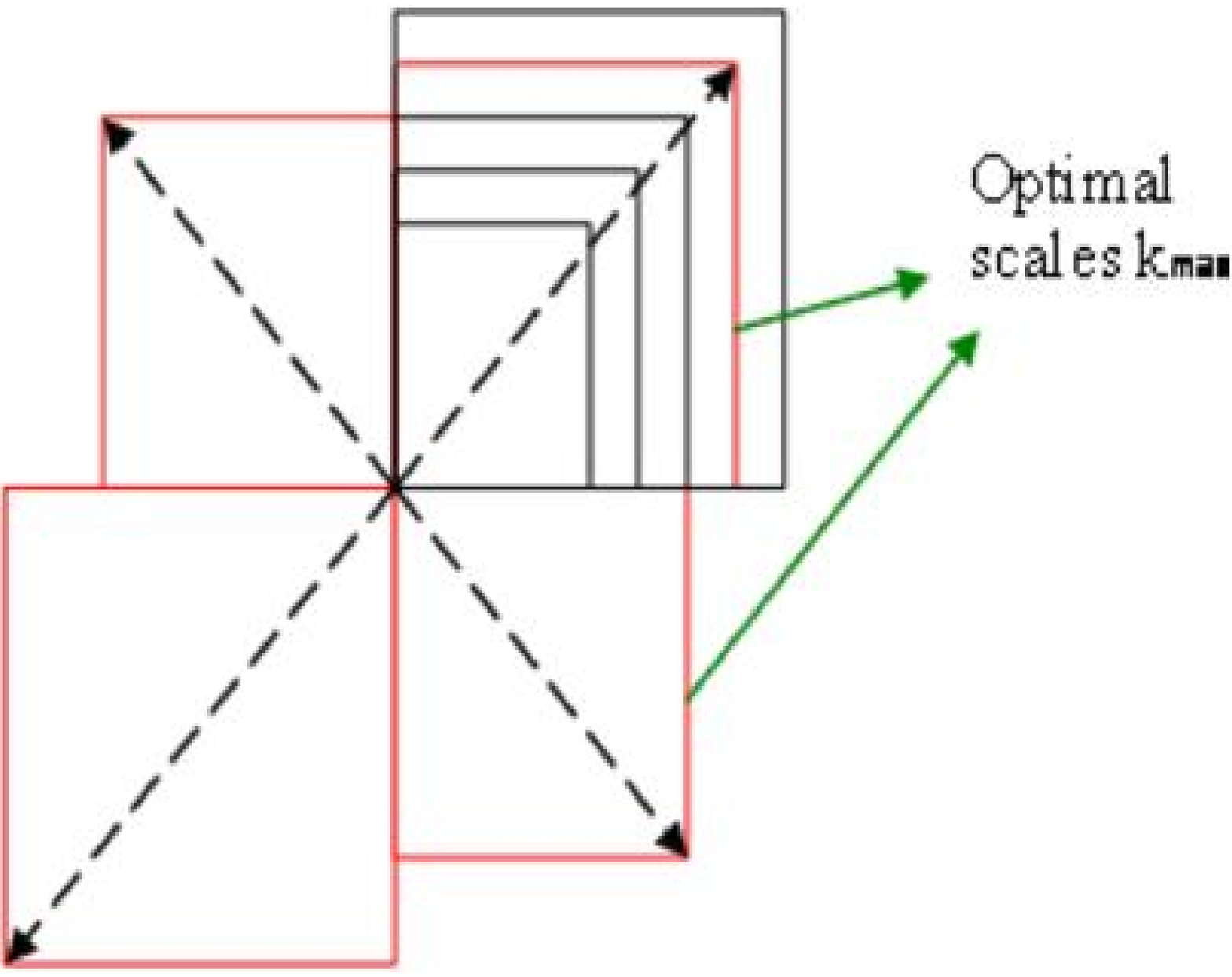}
\label{fig:scale}
}
\subfigure[]{
\includegraphics[width=0.35\linewidth]{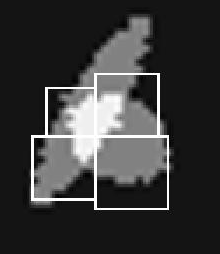}
\label{fig:fourquadrant}
}
\caption{~\ref{fig:scale}: The figure  shows the 4 quadrants for a seed point. It shows the range of scales for entropy calculation for one of the quadrants. The optimum scales for each of the quadrants are shown in red with dotted lines representing the displacement vectors. ~\ref{fig:fourquadrant}: A sample result obtained using the quadrant method on a synthetic image Figure . As can be seen, the four quadrants localize the target salient object.}

\end{figure}

In this algorithm, we initialize a uniformly sampled grid of points on the image. In order to capture all the entropy maxima, we assume that each salient region has at least one grid points within its vicinity. Let us denote the locations of these initial grid points with $\bar{P}_{i}^0$ where the subscript represents the index of the $i^{th}$ point and the superscript denotes the iteration. Each of the points in the grid represents the tentative center of a salient region for the corresponding iteration. At every step, direction of increasing entropy has to be calculated. This is achieved by dividing the neighborhood of the point considered into four quadrants as shown in the figure below. In each of the quadrants we take windows  $W_{jk}(P_{i}^{t})$ of all scales in a range varying over $k$ and compute respective entropies. For each quadrant, the window at the scale that gives maximum entropy is selected. 
The effective shift vector $\bar{\delta ed_{i}}$ is calculated as the summation of the displacement vectors corresponding to the optimal scales weighted by their normalized entropies. For more details see Algorithm~\ref{a:quadrant} and Figure~\ref{fig:scale}.

\vspace{-0.1cm}

\begin{algorithm*}
\caption{Quadrant Method}
\label{a:quadrant}
\begin{algorithmic}
\State Uniformly sampling the image with an initial grid of points $P_{i}^0$ 
\State Initialize magnitude effective displacement vector $\bar{\delta ed_{i}}$ to $\eta$
\For {\textbf{each} point $P_{i}$}
\While{$\|\bar{\delta ed_{i}}\| \leq \eta$}
	\For {\textbf{each} quadrant j centered at $P_{i}$}
		\For {\textbf{each} scale k in the range of scales}
			\State Calculate the entropy $\epsilon_{ijk}$ in window $W_{jk}$ centered at point $P_{i}^{t}$
		\EndFor
		\State $\epsilon_{ij}$ = $max_{k}{\epsilon_{ijk}}$ 
		\Comment{\footnotesize{Find the maximum entropy for each quadrant across the range of scales}}
		\State ${k_{opt}}_{ij}$ = $arg max_{k} \epsilon_{ijk}$
		\Comment{\footnotesize{Find the corresponding scale maximizing the entropy}}
	\EndFor
	\State $ n\epsilon_{ij} = \frac{\epsilon_{ij}}{\sum_{j} \epsilon_{ij}}$ 
	\Comment{\footnotesize{Find the normalized entropy for each quadrant}}
	\State $d_{ij} = \sqrt{2}k\_opt_{ij}$ 
	\Comment{\footnotesize{Displacement in each quadrant direction}}
	\State $\bar{\delta ed_{i}} = \sum_{j}n\epsilon_{ij}\bar{d}_{ij}$
	\Comment{\footnotesize{Calculate effective displacement for the point}}
	\State $t = t + 1$
	\State $\bar{P}_{i}^{t+1} = \bar{P}_{i}^{t} + \bar{\delta ed_{i}}$
	\Comment{\footnotesize{Change location of point}}
\EndWhile
\EndFor
\end{algorithmic}
\end{algorithm*}

Once entropy maxima are found we calculate the change in pdf about the optimal scale. Salient regions are obtained by filtering out the high entropy noise by applying a lower bound on the pdf-difference. A sample result of this approach is shown in Figure ~\ref{fig:fourquadrant}. 
The quadrant approach considers variable sized neighborhoods of a test point and makes an entropy weighted move towards the salient region. The entropy weights are coarse, however, in the sense that they are computed for each quadrant rather than pixel level. In the next section we propose a more fine-grained approach to make more precise shifts to higher entropy regions. While the idea of using quadrants is feasible for 2D problems, its extension to 3D volumes is computationally intensive. Furthermore it cannot deal with 
anisotropic salient regions as described in the next section.


\vspace{-0.3cm}

\subsection{Saliency shift}

Mean shift tracking~\cite{comaniciu} is a popular approach for non-rigid object tracking based on maximizing the Bhattacharyya coefficient between histogram of the target and the candidate window in successive frames. This maximization is achieved by an iterative procedure which involves computing a weight map over the candidate window that reflects the likelihood of a pixel belonging to the target histogram.  A shift vector pointing to the centroid of the weight map is then computed and the procedure is repeated till convergence. We adapt this concept for maximizing entropy. One problem with this is that unlike the visual tracking problem where the target histogram refers to a fixed template, we do not have a specific target histogram to work with for the entropy problem. This can be addressed as follows: We note that among all discrete distributions, the uniform distribution has highest entropy. In order to adapt the meanshift tracking procedure to seek entropy maxima, a simple solution is to use the uniform distribution as the target histogram. We now describe the procedure in detail. A sparse set of seed points is distributed throughout the image. A search window $W_{x}$ is centered around each candidate point $x$ and candidate feature distribution $p$ of the the window is calculated by aggregating the weighted kernel responses over all the pixels in it.

{\footnotesize 
\begin{equation}
p_{b}(W_{x}) = 
		\newline C_{H} \sum_{s \in W} |H|^{-1/2} K(d(x,s))\delta[\beta(s)-b]
\end{equation}
}
where $b$ is the histogram bin index, $\beta(s)$ is the bin number in which pixel $s$ lies, $C_{H}$ is the normalization constant, $H$ denotes the bandwidth matrix, $\delta$ is the discrete Kronecker delta function and $d$ is the euclidean distance. As mentioned above, to move towards higher entropy regions, we define the target distribution to be uniform $q \sim U[0,M]$. The algorithm estimates the shift by maximizing the similarity between the candidate distribution and uniform pdf, measured in terms of the Bhattacharyya coefficient given by $\rho(x) = \sum_{b = 1}^{M} \sqrt{p_{b}(W_{x})q_{b}}$.
Each pixel in the candidate window is assigned weights given by equation ~\ref{eq:weight}

{\footnotesize
\begin{equation}
w(s) = \sum_{1}^{M}{\sqrt{q_{b}/p_{b}(x)}\delta[\beta(s) - b]}
\label{eq:weight}
\end{equation}}

which in effect assigns higher importance to the candidate pixels belonging to the target distribution.
At each iteration, the next position $x$ of the seed point is calculated by a kernel $K$ weighted mean of the candidate window pixels 

{\footnotesize
\begin{equation}
{\textbf{x}} = \frac{\sum_{s \in S}-K'_{H}(x-s)w(s){\textbf{s}}}{\sum_{s \in S}{-K'_{H}(x-s)w(s)}}
\label{eq:pos_estimation}
\end{equation}
}

\begin{figure}[ht]
\centering
\includegraphics[width=0.75\linewidth]{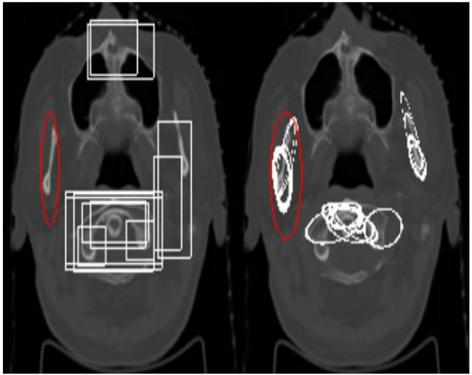}
\caption{The image on the left shows a 2D projection of the output for the saliency detection algorithm using the conventional meanshift framework, whereas the right image shows the same volume slice using the ABMSOD framework. The cheek bones are not detected using conventional meanshift because of the anisotropy.}
\label{fig:abmsodmot}
\end{figure}

The algorithm increasing the Bhattacharyya coefficient with the uniform distribution effectively moves the candidate in a higher entropy direction till the maximum is achieved. All maxima with entropy values above a certain pre-defined threshold are selected and the change in pdf is calculated at the specified scale. Regions with high rate of change of pdf are considered to be salient at that scale. 
We use above mentioned framework for identifying 3D salient regions in medical volumes. Specifically, we use cuboid search windows that are initially distributed randomly throughout the medical volume. Each of these windows has a scale parameter generated uniformly in a range constrained by the size of the volume. The size of the window does not change across iterations i.e the bandwidth matrix $H$ remains the same throughout the meanshift procedure. We use the identity function as our kernel $K(x)=x$. This approach worked well for isotropic salient regions as well as for anisotropic regions aligned with the coordinate axes. However for salient regions which are anisotropic and oblique, the cuboids lack the flexibility to precisely encapsulate the salient region resulting in a higher fraction of the background pixels contributing to the candidate histogram thus reducing its entropy and resulting in a missed detection as can be seen in Figure ~\ref{fig:abmsodmot}. 

To address this problem, we use Adaptive Bandwidth Meanshift for Object Detection (ABMSOD) algorithm ~\cite{ABMSOD}. ABMSOD is a meanshift based iterative algorithm used for 2D object detection in computer vision. It simultaneously estimates the position, scale and orientation of the target object. The initialization step involves randomly  scattering elliptical candidate windows with varying sizes throughout the volume. The iterative step of the algorithm consists of two parts. In the first part, new estimate of the candidate location is calculated using the conventional meanshift framework. In the second part, we estimate the scale and orientation parameters that are encoded within the bandwidth matrix of the candidate window. For details of the implementation see ~\ref{a:ABMSOD}.
 The optimal bandwidth corresponds to the scale and orientation that maximizes the Bhattacharyya coefficient at the current position. ~\cite{abmsot} derives the expression for estimating the optimum bandwidth matrix $H$. 

{\footnotesize
\begin{equation}
H = \frac{\sum_{s \in S}(x-s)(x-s)^{T}w(s)}{\sum_{s \in S}{w(s)}}
\label{eq:opt-H-matrix}
\end{equation}
}
~\cite{pdgc} validates the expression ~\ref{eq:opt-H-matrix} for higher dimensions and uses the ABMSOD framework for localizing 3D structures in medical volumes. For this, the feature histogram of the anatomy to be localized is used as the target distribution. We adopt the same framework for maximizing the entropy using uniform pdf as the target distribution. However the algorithm described in ~\ref{a:ABMSOD} is computationally expensive and is amenable to parallelization using GPUs. The scheme used for parallelization of our algorithm is described in the next subsection. 

\begin{algorithm*}
\caption{Adaptive Bandwidth Meanshift}
\label{a:ABMSOD}
\begin{algorithmic}
\State HQ is the target histogram
\For {\textbf{each} candidate ellipsoid  $W_{i}$ centered at $P_{i}$ with a bandwidth matrix $H_{i}$}
\Comment{\footnotesize Perform iterative search}
\State $iter\_cnt \leftarrow 1$
\State $bhatcf \leftarrow 0$ \Comment{\footnotesize Initialize Bhattacharyya coefficient}
\State $max\_bhatcf \leftarrow 0$ \Comment{\footnotesize Maximum Bhattacharaya coefficient across all iterations}
\State $delta\_bhat\leftarrow THRESHOLD$
\State $x_{opt} \leftarrow P_{i}$ 
\State $H_{opt} \leftarrow H_{i}$
\While {($delta\_bhatcf \leq THRESHOLD$  and $iter\_cnt < MAX\_ITERATIONS$)} \Comment{\footnotesize Termination criteria for the meanshift search}
\For {\textbf{each} voxel $v\in W_{i}$}
			\State $HP(b_{v}) \leftarrow HP(\beta(v)) + c \exp{\frac{-d_{v}}{2}}$ \Comment{\footnotesize $HP$ is the candidate histogram local to each thread, $\beta(v)$ is the bin index for voxel $v$ and $d_{v}$ is the distance of voxel $v$ from the center of the ellipsoid} \label{as:stepa}
\EndFor

\For {\textbf{each} voxel $v\in W_{i}$} 
	\State $w_{v} \leftarrow \sqrt{\frac{HQ(\beta(v))}{HP\beta(v)}}$ \Comment{\footnotesize Weight is computed for each voxel as the ratio of the bin heights of the candidate and target histograms.} \label{as:stepb}	
	\State $\delta_{{x}{v}} \leftarrow -K'{H}(v) w_{v} s_{v}$ \Comment{\footnotesize{$\delta_{{x}{v}}$ denotes the contribution of the voxel $v$ having position $s_{v}$ to the change in the position of the ellipsoid}}
	\State $x_{new} \leftarrow \frac{\sum_{\forall v} \delta_{{x}{v}}}{\sum_{\forall v}-K'{H}(v)w_{v}}$ \Comment{\footnotesize {$x_{new}$ is the new position of candidate ellipsoid and is computed according to Equation~\ref{eq:pos_estimation}}}
	\For {\textbf{each} voxel $v\in W_{i}$}
	\newline \textbf{Repeat} Steps~\ref{as:stepa} to Step~\ref{as:stepb} for the candidate ellipsoid $W_{i}$ centered at $x_{new}$
	\State $\delta_{{H}{v}} \leftarrow (x_{new} - s_{v})(x_{new} - s_{v})^{T}w_{v}$ \Comment{\footnotesize{$\delta_{{H}{v}}$ denotes contribution of a voxel to the H matrix}}
	 \EndFor
	 \State $H_{new} \leftarrow \frac{\sum_{\forall v} \delta_{{H}{v}}}{\sum_{\forall v}w_{v}}$ \Comment{\footnotesize $H_{new}$ is the optimum H matrix at the new position of the candidate ellipsoid and is calculated using ~\ref{eq:opt-H-matrix}}
	 \State $bhatcf \leftarrow \sqrt{(HQ)(HP)}$ 
	 \State $delta\_bhatcf \leftarrow max\_bhatcf - bhatcf$
	 \If {$bhatcf > max\_bhatcf$}
	 		\State $max\_bhatcf \leftarrow bhatcf$
	 		\State $x_{opt} \leftarrow x_{new}$
	 		\State $H_{opt} \leftarrow H_{new}$
	 \EndIf
	 \State $iter\_cnt \leftarrow inter\_cnt + 1$
\EndFor
\EndWhile
\State $P_{i\_opt} \leftarrow x_{new}$
\State $H_{i\_opt} \leftarrow H_{new}$
\EndFor
\end{algorithmic}
\end{algorithm*}

\subsection{Parallelization using GPUs}
In this subsection, we describe the scheme used to parallelize the ABMSOD algorithm on a GPU. A similar scheme is used to accelerate the conventional Meanshift algorithm. The GPU is a data-parallel computing device consisting of a set of multiprocessing units (SM), each of which is a set of SIMD (single instruction multiple data) processing cores. Each SM has a fixed number of registers and a fast on-chip memory that is shared among its SIMD cores. The different SMs share a slower off-chip device memory. Constant memory and texture memory are read-only regions of the device memory and accesses to these regions are cached. Local and global memory refers to read-write regions of the device memory and its accesses are not cached.In the CUDA context, the GPU is called device, whereas the CPU is called host. Kernel refers to an application that is executed on the GPU. A CUDA kernel is launched on the GPU as a grid of thread blocks. A thread block contains a fixed number of threads. A thread block is executed on one of the multiprocessors and multiple thread blocks can be run on the same multiprocessor (For details on GPU architecture see ~\cite{cuda}). The independent exploration of different search paths originating from each initial random point is distributed amongst thread blocks. In addition, using the finer level of parallelism offered by threads within a thread block, we further parallelize operations within each search iteration. We make the threads in a thread block handle computations for a subset of the voxels from the window. Computations such as the application of the kernel function, construction of the candidate histogram, weight assignment to the voxels, H matrix computation etc. are all done in parallel where each thread is responsible for a set of voxels. Summation of values across threads is performed through parallel reduction. To reduce the synchronization operations among threads during histogram computation, we allow each thread to construct a local histogram of the voxels handled by that thread. These histograms are stored in shared memory for fast access. After all local histograms are constructed, the histograms are bin wise aggregated by the threads in parallel to form the global candidate histogram. The volume is stored in a 3D texture and the access to the volume is ensured to be in a way that maximizes spatial locality and efficiently utilizes the texture cache. Constant variables  the target histogram are stored in constant memory to utilize the constant cache. Data shared by threads within a block like the local histograms etc. are stored in shared memory for fast retrieval through the broadcast mechanism supported by CUDA.  Enough number of threads are launched to keep all the cores of each streaming multiprocessor (SM) busy. We try and maximize the occupancy for each SM. The occupancy is however limited  by the amount of shared memory and the number of registers available per SM. We also ensure that there is no register spilling and uncoalesced global memory accesses. 

\vspace{-0.2cm}
\section{Experiments}
We consider the usage of fast salient region detection in order to speed up medical workflows. Specifically, we demonstrate the efficacy of our algorithm in estimating the location and size of left ventricle in myocardial perfusion imaging and quantifying brain tumor in MR images acquired by Fluid Attenuated Inversion Recovery Pulse Sequence (FLAIR). 

\subsection{LV detection in Myocardial perfusion imaging}
\label{PET}

Myocardial perfusion imaging is a nuclear medicine procedure that illustrates the function of the heart muscle (myocardium). The patient is typically administered FDG - fluorodeoxyglucose - which has radioactive isotope fluorine that emits imagable positrons. This technique captures the functional information of the body as against structural information from CT because the glucose part in the radiopharmaceutical rushes to regions in the body such as the myocardium which have high metabolic activity . Physicians require the myocardium dataset in a standard orientation and scale. Current techniques that estimate cardiac orientation ~\cite{comaniciuDatabaseGuided} need a good initial mask around the myocardium otherwise, liver and other nearby organs having high uptake contribute to a biased estimate of the orientation and size. 

Detection of Left ventricle in a medical volume of size $128\times128\times34$ using object appearance based classifier at all possible locations and scales ~\cite{comaniciuDatabaseGuided} is extremely compute intensive. Myocardium in PET has increased uptake value because of high metabolic activity. One way to solve the problem of high computation is using this high uptake principle. Setting a high SUV threshold gives a rough initial estimate of the heart location. But we also observed large number of false positives due to noise, liver and other organs \ref{tab:PET-comparison}. Also, such a simple threshold does not give any estimate of the size in each dimension. Exploring alternatives, we observed that left ventricular region can be considered salient because of the high entropy and uniqueness of its pixel intensity distribution compared to its immediate surroundings. In order to leverage that property, we propose to use 3D saliency seek as a pre-processing step to identify tentative candidates having left ventricle. The candidates would then be further processed for location and cardiac orientation estimation. This 2 step approach eliminates false positives, effectively improving the accuracy and reducing compute time. We apply the saliency seek algorithm on an initial sparse seed point grid distributed uniformly across the volume at multiple initial scales. Once these points converge to local saliency maxima, we pick the top 20 having high pdf difference w.r.t surrounding. We observed that one of the top 20 detections is a true positive in 29 of the 32 volumes dataset giving us a recall rate of $91\%$.

\begin{table}[ht]
\centering
\begin{tabular}{|c|c|c|}
\hline
 Method & Recall Rate & Final precision \\
 				&	(for 20 detections) & (using Hu moments)\\
\hline
 Threshold & 15/32 = 46 \%& N.A$^*$ \\
 Saliency Seek & 29/32 = 91 \%& 25/32 \\
\hline
\end{tabular}
\caption{Comparison of saliency seek vs intensity based thresholding. {\footnotesize $^*$ The intensity thresholding method does not give scale information so it cannot be used for further Hu based detection} }
\label{tab:PET-comparison}
\end{table}

\begin{figure}[ht]
\centering
\includegraphics[width=0.7\linewidth]{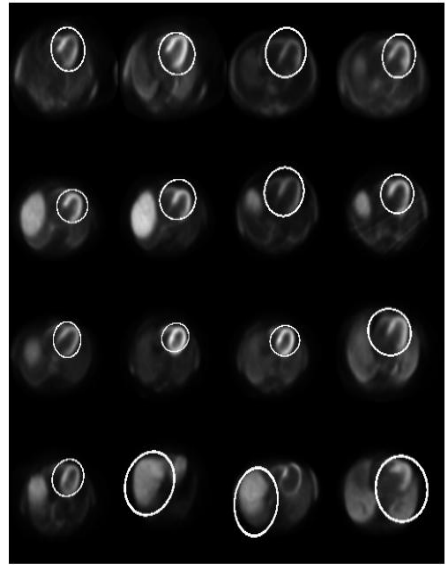}
\caption{Results for LV detection on PET volumes. The figures show 2D axial slices for 16 PET volumes along with the final ellipsoids. The LV is correctly detected in all but two cases.}
\label{fig:PetResults}
\end{figure}

In order to increase the precision, i.e. to identify the true positives among the tentative candidates, we used Hu moments \cite{huPaper}. These central moments have been carefully designed to be invariant to translation, rotation, scale and so they serve as a good choice for describing local appearance. We evaluated the set of 7 invariant Hu moments on the center slice of a heart template forming a 7 dimensional training feature vector. For each test volume, we then compute the 7 dimensional Hu-moment test vectors on five slices about the central slice for each detection. The detection which minimizes the average euclidean distance with the template across the five slices is chosen as the most accurate estimate for that volume. We used a test dataset of 32 PET volumes having a combination of 'rest' and stress' acquisitions each of size $128\times128\times34$. The number of initial seed points was chosen to be 400. Also the average number of iterations for convergence was $~10$. We are able to detect the left ventricle by minimizing the distance in 25 volumes accurately. The average Jaccard index between groundtruths and  successful detections is as high as 41.36\%. See figure ~\ref{fig:PetResults}. Since we have to search for the ventricle structure throughout the PET volume, the algorithm is computationally intensive. To detect the left ventricle in a reasonable amount of time, we use the Nvidia Tesla C2050 GPU as an accelarator to speed up the saliency seek algorithm. This GPU has 14 multi-processors each having 32 cuda cores, resulting in a total of 448 cuda cores. The cores are clocked at 1.15 GHz. Each multi-processor has 48 KB of shared memory and 32 K registers. The GPU device has 3 GB of device memory. With the parallelization scheme as decribed in the previous section, the saliency seek algorithm is able to detect the left ventricle in 4.1 seconds which is 10 times faster as compared to a sequential implementation. 

\vspace{-0.2cm}

\begin{figure}[ht]
\centering
\includegraphics[width=0.9\linewidth]{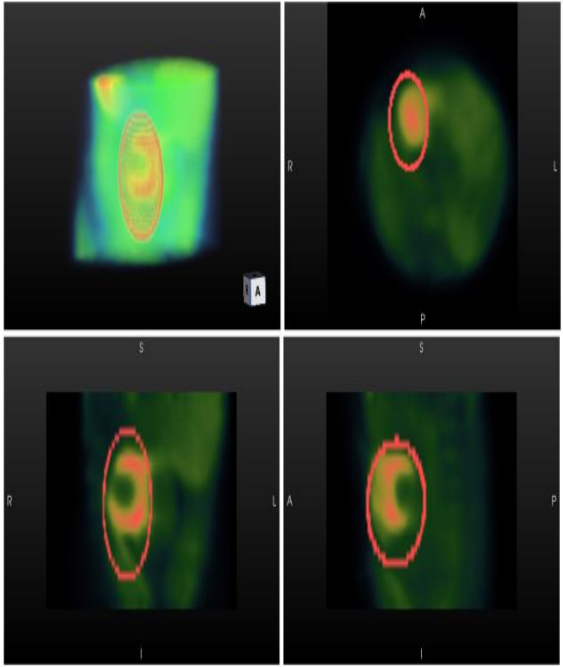}
\caption{The four views (from top to bottom, left to right represent the 3D, axial, coronal and  sagittal views of the PET volumes with LV ventricle detected by the final ellipsoid}
\label{fig:Pet_views}
\end{figure} 

\vspace{-0.3cm}

\subsection{Brain tumor quantification in MR}
In this section we introduce the concept of user defined saliency and demonstrate its accuracy in localizing brain tumor in MR volumes. In brain MRI, the accurate location of tumor and edema is essential for minimizing the damage to healthy tissue. In Fluid Attenuated Inversion Recovery (FLAIR) sequences, even subtle lesions stand out against attenuated csf fluids making this modality conducive for tumor detection i.e. lesions can be distinguished based on intensity values alone and can be considered salient at the right window level setting. Medical images like CT, MR and PET have high dynamic ranges and different tissue types in the body lie in mutually exclusive intensity ranges ~\cite{chungChan} ~\cite{NabilMalej}. For example in CT, lung tissue ranges between -400 and -600 hounsfield units, fat tissue between -60 and -100, soft tissue lies between 40 and 80HU and bone from 400 to 1000 ~\cite{CTintroduction}. In such cases, the range of intensities defined by the window-level setting to be used is determined by anatomy of interest and the application at hand. We consulted a few clinical experts and found out the specific window-level setting used in FLAIR sequences to illustrate malignant lesions. Entropy evaluated over complete dynamic range of MR gave us unwanted regions straddling bone-tissue and air-tissue boundaries. We noticed much more relevant entropy values when the entropy is computed on the constrained intensity range coming from tumor specific window-level. This new entropy quantifies the variation in the distribution of pixels falling in the relevant window-level.

We use the proposed saliency seek algorithm to locate tumors present in such MR volumes. However, we incorporated the new entropy evaluation as discussed above, using only a constrained intensity range. We found empirical evidence showing this idea of application specific saliency improving the detection accuracy significantly. We present the preliminary results of our evaluation of this concept on MR datasets. Brain tumor image data used in this work were obtained from ~\cite{dataBaseMR}. The challenge database ~\cite{dataBaseMR} contains fully anonymized images with manually labeled tumor groundtruth. This dataset consists volumes of size $256\times256\times176$ . We used the saliency seek algorithm in 15 such volumes and were able to successfully localize the tumor in all the 15 volumes with an average Jaccard index of $31.66\%$.

\begin{figure}[ht]
\centering
\includegraphics[width=0.55\linewidth]{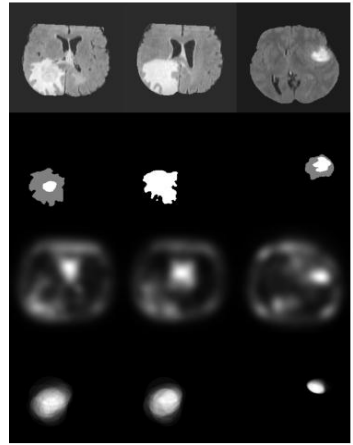}
\caption{The first row shows 2D slices from 3 MR volumes with a tumor. The second row isolates the tumor in each of the volume. Third row shows the output of modified Itti Koch algorithm. The fourth row consists of the set of detections obtained from the saliency seek algorithm.}
\label{fig:MRResults}
\end{figure}

 In order to compare our results with the state of the art we chose to apply a modified version of the Itti Koch approach on axial 2D projections of the brain MR images. The modified algorithm considers only pixels in the constrained intensity range in constructing the saliency maps.  In figure ~\ref{fig:MRResults} the first row shows 2D slices from 3 MR volumes with a tumour. The second row isolates the tumour in each of the volume. The third row consists of the saliency maps obtained using the modified Itti Koch algorithm. The fourth row consists of the set of detections obtained from the saliency seek algorithm. As can be seen from the figure, the Itti Koch algorithm is not able to identify the tumour as a salient region in all the cases, whereas saliency seek employing the metrics of entropy and pdf difference is successfully able localize the tumour.
Since saliency seek in this case consists of searching for the tumor from numerous seed points scattered in a large sized MR volume, the acceleration due to GPUs becomes important for detection in a reasonable amount of time. The number of initial seed points in this case was 700 with the average number of iterations was $~12$. We use the Tesla C2050 GPU as described in ~\ref{PET} and are able to successfully localize the tumor in 7.8 seconds. We obtain a speedup of around 80x over the sequential CPU implementation.
 
\vspace{-0.5cm}

\section{Conclusion}
In this paper, we showed that in the vicinity of a salient region, entropy is a monotonically increasing function of the degree of overlap of a candidate window with the salient region and proposed two iterative approaches to locate salient regions from a sparse grid of seed-points. The first used a four quadrant approach to find entropy maximizing moves. The second used meanshift tracking framework using a uniform target distribution in meanshift iterations for seeking high entropy regions. We developed an efficient GPU implementation of the proposed algorithm for quickly detecting salient regions in 3D and showed promising results for Myocardium detection in PET volumes and tumor quantification in brain MR sequences. The framework can be easily extended for visual tracking by using the converged salient regions from previous frames. Also incorporating target specific feature information within the saliency shift iterations would be another interesting extension.

%
%
%
%

%
%

{\small
\bibliographystyle{ieee}
\bibliography{egbib}
}

\end{document}